\documentclass[conference,dvipsnames]{IEEEtran}
\usepackage{times}
\usepackage{amssymb}
\usepackage{amsmath}
\usepackage[noadjust]{cite}
\usepackage{bm}
\usepackage{listings}
\usepackage{graphicx}
\usepackage[draft]{minted}
\usepackage[colorinlistoftodos]{todonotes}
\usepackage[numbers]{natbib}
\usepackage{multicol}
\usepackage[bookmarks=true]{hyperref}
\usepackage[export]{adjustbox}
\usepackage{xcolor} 
\usepackage{svg}
\newcommand{\blue}[1]{{\normalsize{{\color{black}#1}}}}

\newcommand{\cbfkit}{\textsc{CBFkit}}

\definecolor{codegreen}{rgb}{0,0.6,0}
\definecolor{codegray}{rgb}{0.5,0.5,0.5}
\definecolor{codepurple}{rgb}{0.58,0,0.82}
\definecolor{backcolour}{rgb}{0.95,0.95,0.92}
 
\lstdefinestyle{mystyle}{
    language=Python,
    backgroundcolor=\color{backcolour},   
    commentstyle=\color{codegreen},
    keywordstyle=\color{magenta},
    numberstyle=\tiny\color{codegray},
    stringstyle=\color{codepurple},
    basicstyle=\ttfamily\footnotesize,
    breakatwhitespace=false,
    breaklines=true,
    captionpos=b,
    keepspaces=true,
    numbers=left,
    numbersep=5pt,
    showspaces=false,
    showstringspaces=false,
    showtabs=false,
    tabsize=2,
    basicstyle=\fontsize{6}{7}\selectfont\ttfamily,
}
 
\newboolean{commentsOn}
\setboolean{commentsOn}{False}

\ifthenelse{\boolean{commentsOn}}{\newcommand{\bardh}[1]{{\color{blue}{\bf BH:} #1}}}
{\newcommand{\bardh}[1]{}}

\DeclareMathOperator{\R}{\mathbb R}

\begin{document}

\title{Model Predictive Path Integral Methods with Reach-Avoid Tasks and Control Barrier Functions}

\author{\authorblockN{Hardik Parwana, Mitchell Black, Georgios Fainekos, Bardh Hoxha, Hideki Okamoto, Danil Prokhorov}
\authorblockA{Toyota Motor North America R \& D \\ {firstname.lastname@toyota.com}}}



%

\maketitle

\begin{abstract}
The rapid advancement of robotics necessitates robust tools for developing and testing safe control architectures in dynamic and uncertain environments. Ensuring safety and reliability in robotics, especially in safety-critical applications, is crucial, driving substantial industrial and academic efforts. In this context, we extend \cbfkit{}, a Python/ROS2 toolbox, which now incorporates a planner using reach-avoid specifications as a cost function. This integration with the Model Predictive Path Integral (MPPI) controllers enables the toolbox to satisfy complex tasks while ensuring formal safety guarantees under various sources of uncertainty using Control Barrier Functions (CBFs). \cbfkit{} is optimized for speed using JAX for automatic differentiation and jaxopt for quadratic program solving. The toolbox supports various robotic applications, including autonomous navigation, human-robot interaction, and multi-robot coordination.
The toolbox also offers a comprehensive library of planner, controller, sensor and estimator implementations. 
Through a series of examples, we demonstrate the enhanced capabilities of \cbfkit{} in different robotic scenarios.
\end{abstract}

\IEEEpeerreviewmaketitle

\section{Introduction}

The field of robotics is advancing rapidly, with systems now capable of operating in highly dynamic and uncertain environments. These systems are increasingly deployed in safety-critical applications, where failures can have severe consequences, making safety and reliability paramount. This drives significant industrial investment and academic research focused on developing methods that provide formal safety guarantees, especially for complex, multi-robot scenarios.

Rapid development and testing of new methods are essential in this evolving field. Researchers and developers need tools for rapid prototyping of proof-of-concept ideas to demonstrate the interaction of various control architectures. These architectures typically integrate high-level planning, nominal and feedback control, along with complex sensing and estimation solutions. Efficient integration of these components allows for swift iteration and validation of new approaches, advancing robotic capabilities.

To address these needs, we extend \cbfkit{} \cite{black2024cbfkit}, an open-source Python/ROS2\footnote{\url{https://github.com/bardhh/cbfkit.git}} toolbox designed to facilitate the rapid prototyping and deployment of safe control architectures for robotic systems. Built on Python and using JAX \cite{jax2018github}, \cbfkit{} provides an efficient platform for developing and testing complex autonomy stacks. 
It supports a wide range of applications, including autonomous navigation \cite{BlackFHPP2023icra,YaghoubiEtAl2021cdc}, human-robot interaction \cite{MajdEtAl2021iros,ParwanaEtAl2023arxiv}, multi-robot coordination, and manipulation tasks. 
The toolbox combines model-based and model-free control approaches, offering flexibility to accommodate various system dynamics and control requirements. 
It uses JAX for automatic differentiation and jaxopt for fast quadratic program (QP) solving, resulting in significantly faster computation times compared to symbolic methods. Additionally, the toolbox provides a comprehensive library of implementations for various systems, sensors, estimators, Control Barrier Functions (CBFs), and tutorials for single and multi-agent applications. CBFs have been used to provide safety guarantees in various applications such as (semi-)automated driving~\cite{AmesGT2014cdc,XiaoEtAl2021iccps,HeZZS2021acc,MolnarEtAl2022ifac, cosner2023learning}, arm manipulators~\cite{ShawCortezVD21icra,SingletaryEtAl2022ral,FerragutiEtAl2020ras}, and multi-agent coordination~\cite{ParwanaMP2022cdc,LindemannD2019csl,MehmoodEtl2023jas,EmamEtAl2022tr}. Since CBF controllers are myopic, and safety guarantees are only valid if the controller generates a solution, they benefit greatly from a high-level planner that provides waypoints and attempts to avoid problematic scenarios.

In this paper, we demonstrate how \cbfkit{} can be extended with Model Predictive Path Integral (MPPI)~\cite{williams2016aggressive,williams2018robust,gandhi2021robust} controllers using timed reach-avoid specifications \cite{hoxha2016planning,hashemi2024scaling,lindemann2018control}. Reach-avoid specifications, such as 'reach region $r_1$ within 5 seconds and then reach goal region $r_2$, all while avoiding obstacles,' can serve as a cost function for MPPI. Reach-avoid specifications offer a flexible framework for task planning in robotics, ensuring robots follow specific sequences of actions, maintain safety distances, and achieve goals within specified time windows. This integration is further enhanced by incorporating a CBF-based safety filter, enabling the design of controllers that optimize performance while ensuring safety through formal guarantees provided by CBFs.

We showcase the capabilities of \cbfkit{} through a series of examples. 
These examples highlight the complementary benefits of different components in the autonomy stack. Starting with a robust CBF controller alone, we introduce noise into the dynamics, integrate a Kalman filter, add a planner, and finally incorporate a reach-avoid specification as a cost function into the system. The experiments illustrate the system’s behavior at various stages of complexity, demonstrating the effectiveness and versatility of \cbfkit{}.

{\bf Contribution:}
\begin{itemize}
\item We extend \cbfkit{} for full stack autonomy support and demonstrate this with a novel implementation of the Model Predictive Path Integral (MPPI) planner, which uses timed reach-avoid tasks as a cost function.
\item A demonstration of \cbfkit{} with an MPPI planner combined with a Control Barrier Function (CBF) safety filter to ensure robust and safe navigation.
\item Implementation of a library of CBF controllers, along with various sensors and estimators, designed for dynamic and uncertain environments. 
\end{itemize}

\section{Supported Models and Control Design Problems}

Our goal for \cbfkit{} is to be a rapid development, proof-of-concept tool for the development of autonomy control architectures that, at its lowest level, integrates CBF-based feedback controllers to act as safety filters. \cbfkit{} supports a number of different classes of control-affine models (model of system $\Sigma$):

\begin{enumerate}
\item Deterministic, continuous-time Ordinary Differential Equations (ODE): 
\begin{equation}
\label{sys:ode}
    \dot{x}=f(x)+g(x)u,
\end{equation}
where $x\in \mathcal X\subset \mathbb{R}^{n}$ is the system state,  $u\in \mathcal U \subset \mathbb{R}^{m}$ is the control input, $f:\mathbb{R}^{n} \rightarrow \mathbb{R}^{n}$ and $g:\mathbb{R}^{n} \rightarrow \mathbb{R}^{n \times m}$ are locally Lipschitz functions, and $x(0) \in \mathcal X_0 \subseteq \mathcal X$ is the initial state of the system.

\item Continuous-time ODE under bounded disturbances:
\begin{equation}
\label{sys:ode_uncertain}
\dot{x}=f(x)+g(x)u+Mw,
 \end{equation}
where $w\in \mathcal W$ is the disturbance input,  $\mathcal W$ is a hypercube in $\mathbb{R}^{l}$, and $M$ is a $n\times l$ zero-one matrix with at most one non-zero element in each row.

\item Stochastic differential equations (SDE):
\begin{equation}
\label{sys:sde}
 \mathrm{d}{x} = \big(f({x}) + g({x}){u}\big)\mathrm{d}t + \sigma({x})\mathrm{d}{w}
 \end{equation}
where $\sigma: \R^n \rightarrow \R^{n \times q}$ is locally Lipschitz, and bounded on $\mathcal X$, and ${w} \in \R^q$ is a standard $q$-dimensional Wiener process (i.e., Brownian motion) defined over the complete probability space $(\Omega, \mathcal{F}, P)$ for sample space $\Omega$, $\sigma$-algebra $\mathcal{F}$ over $\Omega$, and probability measure $P: \mathcal{F} \rightarrow [0,1]$. 

\end{enumerate}


In certain practical applications, not all the states of the system may be observable. 
In such scenarios, we may assume that a state vector $y$ is observable.
For example, in the case of SDE, we may assume:
\[ dy = Cx \mathrm{d}t + D \mathrm{d}v \]
where $C \in \mathbb R^{p \times n}$, $D \in \mathbb R^{p \times r}$, and $v \in \R^r$ is a standard Wiener process.

\section{Autonomy Stack}

\begin{figure}[h!]
    \centering
    \includegraphics[trim=0 35 0 35, clip, width=200px]{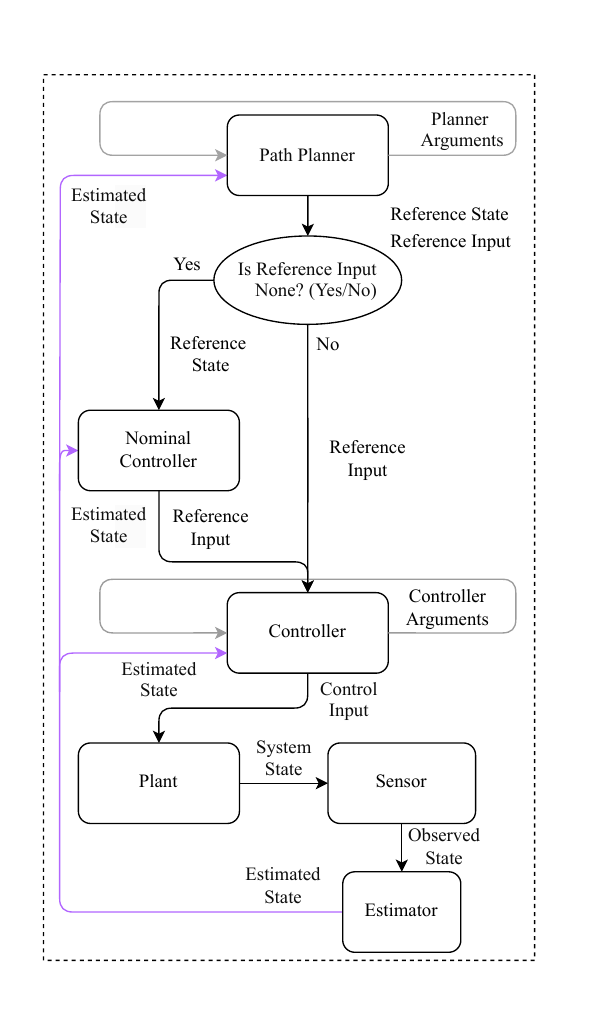}
    \caption{Closed-Loop Simulator in \cbfkit{}.}
    \label{fig:closed-loop}
    \vspace{-15pt}
\end{figure}

In this section, we describe the architecture of an autonomy stack built using \cbfkit{}. 
An autonomy stack typically comprises of sensors, estimators, planners and controllers, each responsible for different aspects of autonomous operation. 
As shown in Fig. \ref{fig:closed-loop}, we outline a three-tier architecture consisting of a high-level planner, a nominal controller, and a feedback controller. 
This structure ensures that the robotic system can navigate complex environments while maintaining safety and robustness.

\subsubsection{High-Level Planner}

The high-level planner generates a feasible path from a starting point to a goal location, considering global objectives and constraints such as avoiding large obstacles and navigating dynamic environments. 
In \cbfkit{}, the Model Predictive Path Integral (MPPI) controller is used as the high-level planner. 

\blue{We provide a brief description of the MPPI algorithm. MPPI is a sampling-based procedure to solve a finite horizon model predictive control problem.
Consider a nonlinear system with state $x_t \in \mathbb{R}^{n}$ and control input $u_t \in \mathbb{R}^m$ that follows the following discrete-time dynamics}
\begin{align}
   x_{t+1} = F(x_t, u_t). 
\end{align}

\blue{For a time horizon $H$, consider the state trajectory $\mathbf{x}=[x_t^T, ..., x_{t+H}^T]^T$, mean control input sequence $\mathbf{v} = [v_t^T, .., v_{t+H}^T]^T, v_\tau\in \mathbb{R}^m$ and injected Gaussian noise $\mathbf{w} = [w_t^T,..,w_{t+H}^T]^T$ where $w_\tau \sim N(0,\Sigma_w)$ where $\Sigma_w$ is the noise covariance, often chosen by the user. Let the disturbed control input sequence be $\mathbf{u}=[u_t,...,u_{t+H}]=\mathbf{v}+\bm{w}$.  
MPPI then solves the following problem}
\begin{subequations}
    \begin{align}
        \min_{\mathbf{v}} \quad  & J(\mathbf{v}) = \mathbb{E} \Biggl[ Q(x_t,..x_{t+H},\mathbf{u}) +  \nonumber \\ 
 & \quad\quad\quad\quad\quad\quad \sum_{\tau=t}^{t+H-1} \left( \frac{\lambda}{2} v_\tau^T \Sigma_w^{-1} v_\tau \right)  \Biggr] \label{eq::mppi_objective_org}\\
   \textrm{s.t.} \quad & x_{\tau+1} = F(x_\tau, v_\tau+w_\tau) \\
   & w_\tau \sim \mathcal{N}(0,\Sigma_w) \label{eq::MPPI-disturbance}
\end{align}
\label{eq::mppi_objective}
\end{subequations}
\blue{More algorithmic details to solving \eqref{eq::mppi_objective} can be found in \cbfkit{} as well as MPPI papers\cite{williams2016aggressive,williams2018robust,gandhi2021robust}. 
Next, we present some details on $Q$ in \eqref{eq::mppi_objective_org}.
}

\blue{The cost functions are generally user-designed for specific objectives. For instance, we introduce cost metrics here to quantify progress towards achieving two types of user-specified tasks: convergence to a goal and collision avoidance.

Convergence to a goal $g$ within a radius $r_g$ is specified in terms of distance to the goal using cost $c_g$ as follows:
\begin{align}
    c_g(x_t) = k_g (|| p_t - p_{g,t} ||^2 - r_g^2)
    \label{eq::stl_goal}
\end{align}
where $p_{g,t}$ is the location of the goal at time $t$, $k_g>0$ is a weighting factor, and $p_t$ is the location of the robot extracted from its state $x_t$. 
For collision with an obstacle $o$, we use the inverse of distance to the obstacle to define the cost $c_o$ as follows:
\begin{align}
    c_o(x_t) = \frac{k_o}{ \max( ||p_t-p_o||, \epsilon ) }
\end{align}
where $p_o$ is the location of the obstacle and \( 0 < \epsilon \ll 1 \) prevents the cost from becoming infinitely large.
}
\blue{For time-invariant tasks, we design cost functions is as follows
\begin{align}
    Q = \sum_{\tau=t}^{t+H} c_g(x_\tau) + c_o(x_\tau)
    \label{eq::mppi_standard_cost}
\end{align}

For time-dependent tasks such as timed reach-avoid specifications, our MPPI leverages signal temporal logic-inspired robustness metrics to define the cost functions. The timed reach-avoid specifications can specify task requirements, such as reaching a destination within a specified time or avoiding certain regions. The cost functions for reaching the goal $g$ between times $t_1$ and $t_2$ is designed as follows
\begin{align}
    Q_g[t_1, t_2] = \min(c_g(x_{t_1}),..., c_g(x_{t_2}) )
    \label{eq::stl_trajectory_goal}
\end{align}
where $t_1$ and $t_2$ are global times. In \eqref{eq::stl_trajectory_goal}, if $t_1<t$ (or $t_2<t$), $x_{t_1}$ (or $x_{t_2}$), where $t$ is the current time, is obtained from the history of states. 
On the other hand, if $t_2>t_1\geq t$, $x_{t_1}, x_{t_2}$ are obtained using the predicted states during the sampling procedure of MPPI. 
Further note that in contrast to the standard way of designing MPPI costs in \cite{williams2016aggressive}, we allow MPPI cost to also depend on states $x_\tau$ for $\tau<t$ so that a task already achieved in the past is not revisited when MPPI planner/controller is implemented in a receding horizon fashion. 

Similarly, for collision avoidance tasks that should be satisfied for all time $t>0$, we design
\begin{align}
    Q_o = -\max ( c_o(x_o), ... , c_o(x_{t+T}) )
    \label{eq::stl_trajectory_obstacle}
\end{align}
The final cost for a sampled trajectory is designed as
\begin{align}
    Q(x_t,..,x_{t+T}) = \min( Q_g[t_1,t_2], Q_c )
\end{align}
Note that our cost is sensitive to chosen weights $k_g,k_o$ and thus requires some manual tuning. Automated tuning of these hyperparameters will be supported in future library releases. 
Finally, note that the above costs can be compounded for any number of goals and obstacles. 

We would also like to mention that other implementations of MPPI with spatio-temporal specifications \cite{Varnai2022} and CBF shielding have been developed in the past\cite{yin2023shield}.
}

\subsubsection{Nominal Controller}

The nominal controller operates at an intermediate level, refining the plan generated by the MPPI into a sequence of control commands executable by the feedback controller. 
This layer ensures that the planned path is followed accurately, adjusting for any deviations due to model inaccuracies or unforeseen obstacles.

In \cbfkit{}, several controllers have been developed for different systems. 
These include a proportional controller for the bicycle model, a Lyapunov controller for the van der Pol oscillator, and a geometric controller for a quadrotor (6 DOF).

\subsubsection{Feedback Controller}

\begin{table*}[t]
\centering
\begin{tabular}{|l|c|c|c|c|c|}
\hline
\textbf{Controller Type} & \textbf{Vanilla} & \textbf{Robust} & \textbf{Stochastic} & \textbf{Risk-Aware} & \textbf{Use Cases} \\ \hline
CBF & \checkmark & \checkmark & \checkmark & \checkmark & Collision avoidance, navigation \\ \hline
CLF & \checkmark & \checkmark & \checkmark & \checkmark & Path following, tracking \\ \hline
CBF and CLF & \checkmark & \checkmark & \checkmark & \checkmark & Dynamic environments, mixed objectives \\ \hline
\end{tabular}
\caption{Feedback Controllers in \cbfkit{} and their Supported Features}
\label{table:low_level_controllers}
\vspace{-20pt}
\end{table*}

The feedback controller executes the control commands generated by the nominal controller. 
In \cbfkit{}, the feedback controller is implemented using various types of Control Barrier Functions (CBF) and Control Lyapunov Functions (CLF), tailored to different robotic systems, environments, and sources of uncertainty. 
The framework of CBF-based control can provide safety guarantees by enforcing a forward invariant set.
Namely, if the system starts in a safe state, then it should always stay in the safe set.

The CBF controller adjusts the control inputs to ensure the robot remains within safe operating conditions. 
\blue{For the sake of completeness, we provide a short and informal description of vanilla CBFs here. A safe set $\mathcal{S}$ of safe states is defined as the 0-superlevel set of a continuously differentiable function $h(x):\mathcal{X} \rightarrow \mathbb{R}$ as follows:
\begin{align}
    \mathcal{S} & \triangleq \{ x \in \mathcal{X} : h(x) \geq 0 \}, \\
        \partial \mathcal{S} & \triangleq \{ x\in \mathcal{X}: h(x)=0 \}, \\
        \textrm{Int} (\mathcal{S}) & \triangleq \{ x \in \mathcal{X}: h(x)>0  \}.
\label{eq::safeset}
\end{align}
        
The following \textit{CBF condition} is then imposed in a controller
\begin{align}
 \dot h(x,u) = \frac{\partial h}{\partial x}( f(x) + g(x)u ) \geq -\nu(h(x)), ~ \forall x \in \mathcal{X}. 
 \label{eq::cbf_derivative}   
\end{align}
where $f, g$ are functions defining control-affine dynamics in \eqref{sys:ode}.
The condition \eqref{eq::cbf_derivative} essentially restricts the rate at which the robot is allowed to approach the boundary of the safe set. And on the boundary where $h=0$, it pushes the robot back as $\dot h(x,u)\geq 0$. The reader is referred to \cite{ames2016control} for more details. }

By utilizing JAX for automatic differentiation and jaxopt for efficient quadratic program solving, \cbfkit{} enables fast and accurate computation of control inputs, making it suitable for prototype implementations. 
One of the main strengths of \cbfkit{} is its library of various implementations of feedback controllers (see Table \ref{table:low_level_controllers}). 
These include vanilla CBF and CLF controllers, robust CBF and CLF controllers, risk-aware stochastic CBF and CLF controllers, and combined risk-aware stochastic CBF and Lyapunov controllers.

\subsection{Auto-Differentiation for CBF Implementations}

A unique feature of the toolkit is the use of JAX~\cite{jax2018github} for auto-differentiation of the barrier function. 
For complex dynamics, this can be computationally challenging using symbolic toolboxes like SymPy~\cite{10.7717/peerj-cs.103}.

JAX computes derivatives without manually or symbolically differentiating the function, enabling our tool to support arbitrary systems and barrier functions, provided that the barrier functions used for control have relative-degree\footnote{A function $p: \mathbb R_+ \times \mathbb R^n \rightarrow \mathbb R$ is said to be of relative-degree $r$ with respect to the dynamics \eqref{sys:ode} if $r$ is the number of times $p$ must be differentiated before one of the control inputs $u$ appears explicitly.} one with respect to the system dynamics.

For barrier functions with a relative-degree greater than one, our \texttt{rectify-relative-degree} module can derive a new barrier function whose zero super-level set is a subset of that of the original barrier function. 
This is done by iteratively differentiating the original barrier function with respect to the system dynamics until the control input appears explicitly (determined by evaluating samples of the term $\frac{\partial h(x_s)}{\partial x}g(x_s)u$ for samples $x_s \in \mathbb \mathcal \mathcal X$), and applying exponential CBF~\cite{nguyen2016exponential} or high-order CBF~\cite{xiao2019control} principles to return a “rectified” barrier function.

In \cbfkit{}, we provide solutions (feedback controllers) to the above two problems using Quadratic Program formulations as in, e.g., \cite{BlackJSP2023acc,AmesGT2014cdc} for model (\ref{sys:ode}), \cite{YaghoubiFS2020itsc,jankovic2018robust} for model (\ref{sys:ode_uncertain}), and \cite{YaghoubiEtAl2021csl,YaghoubiEtAl2021cdc,BlackFHPP2023icra} for model (\ref{sys:sde}).

\subsection{Closing the Loop with Sensors and Estimators}

In addition to the autonomy stack, the closed-loop system also includes sensors and estimates into one framework. 
Sensor models are integrated into the toolbox to simulate realistic scenarios. 
Estimators are  used to infer the system's state when direct measurements are not available or are noisy. In \cbfkit{} the following estimators are included: extended Kalman Filter (EKF), unscented Kalman Filter (UKF), hybrid EKF-UKF filter. 
We note that all of these are auto-generated based on the systems dynamics equations. For more details on the auto-generation process see \cite{black2024cbfkit}.

The closed-loop simulator also supports a number of integrators, such as forward Euler and solve ivp from SciPy and several integrators from JAX~\cite{jax2018github}.

\section{Simulation Examples}
\label{sec:SimulationExamples}

We provide two simulation studies showing application of our planners and controllers.

\begin{figure}[h]
    \centering
    \includegraphics[width=0.90\columnwidth]{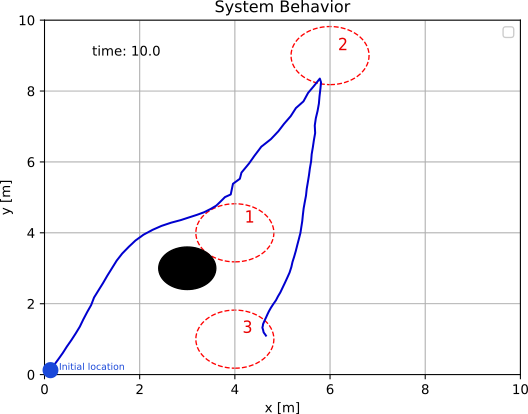}
    \caption{A single integrator robot completes the timed-reach-avoid task 'reach $g_1$ between $0-3.5s$, $g_2$ between $3.6-5s$, and $g_3$ between $5.1-10s$ while avoiding obstacles'. For animation, see here: \href{https://youtu.be/Iie4pq_zVfA}{\texttt{https://youtu.be/Iie4pq\_zVfA}}.}
    \label{fig:mppi_reach_avoid}
    \vspace{-15pt}
\end{figure}

\subsection{MPPI with Timed Reach-Avoid Tasks}
\blue{We use our timed reach-avoid specifications to guide the robot to three waypoints within user-specified time intervals while performing collision avoidance. 
For each goal $i, i\in \{1,2,3\}$, we consider the cost function in \eqref{eq::stl_goal}
We design a time-varying MPPI cost as in \eqref{eq::stl_trajectory_goal}
to promote reaching $g_1$ between 0-3.5s, $g_2$ between 3.6-5s, and $g_3$ between 5.1-10s. 
We also perform collision avoidance with an obstacle (shown in black in Fig.\ref{fig:mppi_reach_avoid}) with cost defined in \eqref{eq::stl_trajectory_obstacle}.
The robot is modeled as a single integrator and the MPPI controller is implemented with a horizon of 50 time steps and 10,000 samples.  
The results are shown in Fig. \ref{fig:mppi_reach_avoid}. 
We see that the robot touches the circle and immediately moves on to the next waypoint. 
}

\subsection{MPPI-CBF controller}

\begin{figure}[h]
    \centering
    \includegraphics[width=0.90\columnwidth]{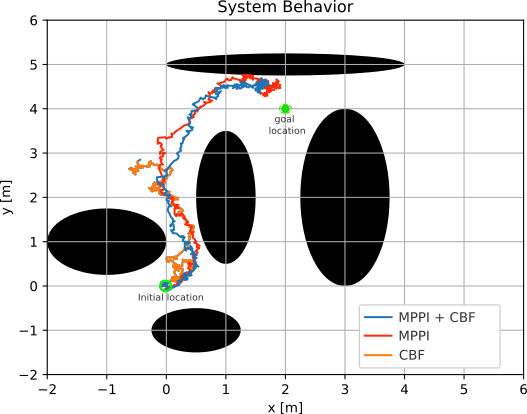}
    \caption{Comparison of three controller setups for navigating from the initial location to the goal location. 
    The stochastic CBF controller (orange) becomes infeasible due to the number of obstacles. 
    The MPPI controller (red) alone violates safety constraints. 
    Finally, the MPPI and stochastic CBF (blue) successfully navigate from the initial to the goal location.}
    \label{fig:scbf-mppi-comparison}
    \vspace{-10px}
\end{figure}

\begin{figure}[h]
    \centering
    \includegraphics[width=0.90\columnwidth]{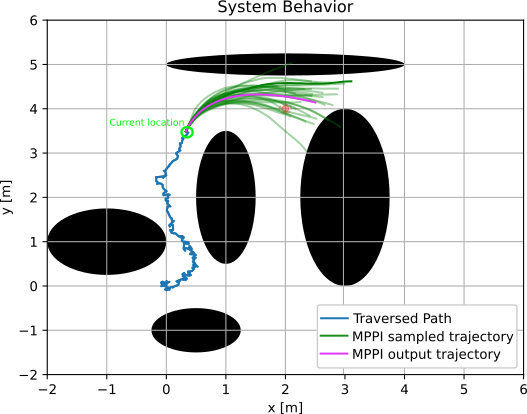}
    \caption{Visualizing MPPI sampled and output trajectories for comparison.}
    \label{fig:scbf-mppi-sampeled}
    \vspace{-10px}
\end{figure}

Consider the scenario shown in Fig. \ref{fig:scbf-mppi-comparison}. 
The objective of the robot is to reach its goal location while avoiding obstacles. 
The robot follows the SDE dynamics in \eqref{sys:sde} with $f(x),g(x)$ defining an extended unicycle model with inputs linear acceleration and angular velocity and a constant noise term $\sigma(x)=0.28$. 
We compare the following three methods: 1) Stochastic CBF (SCBF) QP, 2) MPPI, and 3) MPPI + SCBF. In MPPI + CBF, the MPPI acts as a local planner whose output is filtered by the SCBF QP controller. 
The MPPI in all scenarios is implemented using only the nominal deterministic dynamics. \blue{The MPPI cost are designed as in \eqref{eq::mppi_standard_cost}.}
This is not uncommon in practice as planners typically use simplified dynamics and controllers consider the full dynamics model. 
As such, owing to imperfect dynamics and the non-existence of guarantees of hard constraint satisfaction in theory, MPPI is expected to violate safety constraints.  
We simulate for 5s and the resulting trajectories are visualized in Fig. \ref{fig:scbf-mppi-comparison}. 
The MPPI planner uses a horizon of 80-time steps and 20,000 samples. 
The SCBF QP controller ensures the robot's safety but is unable to get close to the goal. We attribute this to its greedy local optimization from only considering the instantaneous state.
The MPPI controller can get close to the goal however it also gets close to the obstacles and is also observed to collide with the obstacle at the top. 
The MPPI-SCBF performs best and avoids all the obstacles. 
The MPPI can guide the robot in the correct direction owing to its finite horizon planning and other SCBF filters correctly filter its output to provide safety guarantees. To help understand the MPPI execution, we also show a snapshot of the simulation at t=3s in Fig. \ref{fig:scbf-mppi-sampeled}. The MPPI sampled trajectories are shown in green and each sampled trajectory is weighted using our designed cost function. The final output trajectory is computed based on these weights and is shown in pink.

\section{Conclusion} 
\label{sec:conclusion}

The paper presented an extended version of \cbfkit{}, integrating Model Predictive Path Integral (MPPI) methods with reach-avoid tasks and Control Barrier Functions (CBFs) to enhance the safety and robustness of autonomous robotic systems. 
The integration of timed reach-avoid specifications with MPPI provides a powerful framework for task planning and control, enabling robots to navigate complex environments while adhering to safety requirements.


\bibliographystyle{plainnat}
\bibliography{main}

\end{document}